# Autism Spectrum Disorder Screening Using Discriminative Brain Sub-Networks: An Entropic Approach

Mohammad Amin, Farshad Safaei

Faculty of Computer Science and Engineering, Shahid Beheshti University G.C., Evin 1983963113, Tehran, Iran
m.amin.ce.sbu@gmail.com, f_safaei@sbu.ac.ir

**Abstract:** Autism is one of the most important neurological disorders which leads to problems in a person's social interactions. Improvement of brain imaging technologies and techniques help us to build brain structural and functional networks. Finding networks topology pattern in each of the groups (autism and healthy control) can aid us to achieve an autism disorder screening model. In the present study, we have utilized the genetic algorithm to extract a discriminative sub-network that represents differences between two groups better. In the fitness evaluation phase, for each sub-network, a machine learning model was trained using various entropy features of the sub-network and its performance was measured. Proper model performance implies extracting a good discriminative sub-network. Network entropies can be used as network topological descriptors. The evaluation results indicate the acceptable performance of the proposed screening method based on extracted discriminative sub-networks and the machine learning models succeeded in obtaining a maximum accuracy of 73.1% in structural networks of the UCLA dataset, 82.2% in functional networks of the UCLA dataset, and 66.1% in functional networks of ABIDE datasets.

## 1- Introduction

Neurological disorders are structural, biochemical, or electrical anomalies that cause symptoms in the affected person. Alzheimer, Parkinson, and Autism are among the most significant of these disorders [1-3]. Autism spectrum disorder leads to problems in a person's social interactions and communications [4]. One of the strategies for investigating the brain system is to identify its components and analyze functional and structural relations between these regions.

The functional and structural networks of the brain are obtained using a variety of brain imaging techniques such as electroencephalography (EEG) [5], magneto encephalography (MEG) [6], functional magnetic response imaging (FMRI) [7, 8], and diffusion tensor imaging (DTI) [9].

In the structural network, a neuron, a set of neurons, or a region of the brain known as an anatomical parcellation can be a node and physical connections between these nodes, such as synaptic connections or fibrous pathways, are known as edges. Due to the approximate stability of physical connections, the network usually does not change significantly during the imaging process.

In functional networks, the nodes are predefined regions and signal recording begins from them. These regions are named regions of interest. Connections between these regions have dynamic interactive patterns that are obtained by calculating the correlation of recorded time-series data [10].

Differences in some functional and structural connections of the brain regions are usually signs of disease, and the pattern of these changes can be used as biomarkers. Previous studies show that the functional and structural connections of autistic people are different from healthy people [11].

In the current study, for each type of functional and structural network, a sub-network is found using the genetic algorithm which has significant topology differences in two groups and the classification task for autism screening is done using this sub-network instead of the whole network. To fitness evaluation of each sub-network, we need to compare the topological features of the sub-network in two groups so we used various graph entropies to describe network structural characteristics.

The entropy of a graph is a functional depending both on the graph itself and on a probability distribution on its vertex set. This graph functional originated from the problem of source coding in information theory and was introduced in 1973. Although the notion of graph entropy has its roots in information theory, it was proved to be closely related to some classical and frequently studied graph theoretic concepts. The graph entropies are efficient and effective measures for structural characterizing rather than classical graph features [43].

Most previous studies on network classification have used complete networks for their classification. There are logical reasons to choose sub-networks to classify instead of complete networks. A number of them are as follows:
- When using complete network we usually use the edges as features that causes over-fit because of too many features and too little scans.
- Specific sub-networks of the brain can give us more info on different disorders and lower the risk of neutralizing the features of a disorder via some nodes.
- Screening accuracy improves while processing sub-networks.
- Better interpretation can be achieved when we have a discriminative sub-network.

In the current study, some major contributions are:
1. Utilizing various graph entropies for describing brain network topological features.
2. Employing the genetic algorithm for extracting a discriminative sub-network for screening in both types of functional and structural networks.
3. Improving screening accuracy and time.

In Section 2, the related researches on autism screening and graph classification will be studied. In Section 3, the proposed method, and in Section 4 the simulation results will be discussed.

## 2- Related Work

In recent years, many researchers have studied in the field of graph classification and many methods are designed and proposed. The review of recent researches indicates that researchers are either trying to propose criteria to measure the similarities between two graphs [12-17] or new effective features in complex networks to improve machine learning models performance [18-23].

Rudie et al. [11] have studied the differences between functional and structural brain networks and have evaluated these networks in terms of various criteria. By analyzing the functional network of healthy and autistic people, they have found that brain networks of autistic people have less functional integrity and less strong connections between different regions. On the other hand, they have concluded that in some regions the clustering coefficient is decreased, while the paths between different regions have become shorter. Regarding the dissimilarity in structural networks between the two groups, they concluded that in affected people, the integrity of the white matter decreased, but the number of fibers between the regions increased. An analysis of the principal component (PCA), which combines the characteristics of structural and functional networks, has shown that the balance of local and global efficiency is reduced in affected people and this problem has a positive correlation with age and negative correlation with the severity of the symptoms.

Tolan et al. [24] have proposed a method based on the brain networks classification for autism screening. They achieved 67% accuracy in autism detection for functional networks and 68% accuracy for structural networks employing ensemble machine learning models.

Dodero et al. [25] have presented a kernel-based method for network classification using the adjacency matrix of brain networks. Since the adjacency matrix of the brain network is a symmetric positive definite (SPD) and has a Riemannian structure, it is possible to design a new kernel function for support vector machine. The proposed kernel function is designed using Log-Euclidean distance.

A feature-based method for autism disorder screening is proposed by Petro et al. [26]. They used 16 features including edge weights-related and topological features to train machine learning models. Finally, after evaluating the machine-learning model on structural brain networks, they reported an accuracy of 64% for autism screening.

Heinsfeld et al. [27] have presented an autism screening method based on the deep learning models. They have used the brain network adjacency matrix elements as features. Due to the symmetry of the adjacency matrix, the upper triangle and the main diameter of the matrix have been removed, and then the lower triangle values have been normalized and used to train deep networks with various configurations. They evaluated their proposed method with various datasets and obtained 70% accuracy in classification.

Ktena et al. [28] have proposed a metric-based method and have utilized Graph Convolutional Network (GCN) for autism disorder screening. They proposed a novel metric learning method to evaluate the distance between graphs that leverages the power of convolutional neural networks while exploiting concepts from spectral graph theory to allow these operations on irregular graphs. Experimental results on the ABIDE dataset show that their method can learn a graph similarity metric tailored for a clinical application, improving the performance of a simple k-nn classifier by 11.9% compared to a traditional distance metric.

Ataei et al. [29] have presented a feature-based network classification method for autism disorder screening using topology and graphlet frequency features. Their evaluation results indicate the acceptable performance of the screening.

Dodonova and et al. [30] have classified the brain networks for autism screening using the graph spectrum theory features and obtained appropriate detection accuracy.

Eslami and et al. [31] have proposed a new method for autism disorder screening. They have generated brain networks using the SMOTE algorithm and a multi-layered perceptron neural network is trained using this data. Finally, they used the weights of the hidden layers of the neural network to learn some combined models of SVM obtained by the ATM method. Their proposed method has obtained an accuracy of 72.7 to detect autism in the UCLA dataset.

Song et al. [32] have tried to analyze the characteristics of associations in the brain networks for autism screening.

Payabvash et al. [33] have applied machine-learning models such as random forest, support vector machine, neural networks, etc., using features related to the density of edges in the structural brain networks. In addition to gaining an accurate screening model, they have found that the density of the edges in many regions of the brain in autistic people is significantly lower than in people with healthy controls.

Brown et al. [34] have employed a method based on linear regression to obtain the brain sub-network and used this sub-network to autism screening. To take advantage of sub-networks leads to improving machine learning models accuracy and increase the interpretability of results. Their method has achieved a screening accuracy of 66%.

Li et al. [44] have proposed a new network regularized support vector machines method to identify the faulty sub-networks associated with autistic people using diffusion tensor imaging. After constructing the brain connectivity network of each subject using DTI, they utilized the SVM-recursive feature elimination algorithm to identify the faulty sub-networks in order to distinguish ASD from typical developing controls. Since connections in the network are not independent of each other, their topological proximities are incorporated as the network regularization of SVM-RFE by using the graph Laplacian to obtain robust sub-networks. Experiments on both simulated and clinical datasets have shown a better performance of their proposed method in faulty sub-network identification, compared with the traditional SVM-RFE method.

Genetic Algorithm (GA) is an optimization method based on the mechanics of natural genetics and natural selection. Genetic Algorithm mimics the principle of natural genetics and natural selection to constitute search and optimization procedures. GA is used for scheduling to find the near to optimum solution in short time. Since extracting a discriminative sub-network is a time-consuming optimization problem, in the current study we have utilized GA to solve the issue.

Although valuable efforts have been made to screen the autism disorder, the proposed methods do not have acceptable performance, especially in multi-site datasets. In this study, for the first time, the evolutionary algorithms have been used to find the brain sub-networks with significant differences in the network topology in two groups, and graph entropies have been used to describe the topological features of these sub-networks.

## 3- Proposed Method

The focus of the current study is to provide a brain network classification method for autism disorder screening. We have conducted the task using a discriminative sub-network instead of the whole network for the sake of improving the accuracy of the screening models and increasing the interpretability of the results. In the proposed method, we have extracted the discriminative sub-network using the genetic evolutionary algorithm and network entropy features. The main process of the proposed method can be seen in Figure 1.

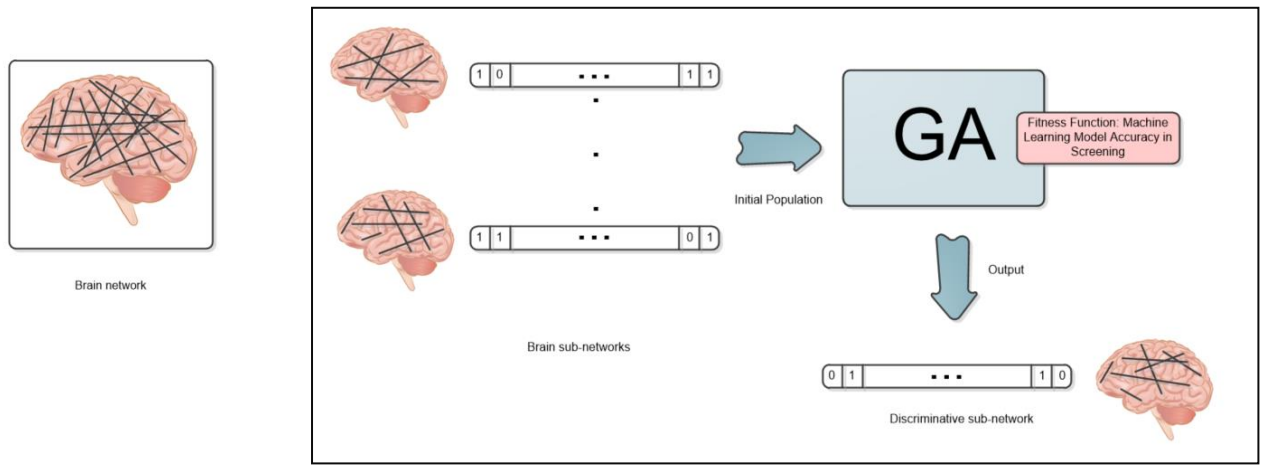

Figure 1: The main process of the proposed method

- **UCLA Dataset**

Resting-state functional magnetic resonance imaging (RS_FMRI) and diffusion tensor imaging (DTI) has been performed using the Siemens T Trio device at UCLA University. The captured images belong to people with standard brain structures, In addition, the brain imaging of people who have metal objects in their bodies has not been performed. The verbal-linguistic and functional intelligence and IQ of the dataset people have been evaluated utilizing Wechsler's criteria. Experts according to the diagnostic and statistical manual of mental disorders (Fifth Edition) [11] has diagnosed autistic people. The UCLA dataset uses an atlas called power-264, developed in 2011 to brain functional mapping by Power and et al. [35]. In this atlas, the human brain is divided into 264 regions.

The UCLA dataset includes 42 functional connection matrices for affected people and 37 functional connection matrices for healthy people. The structural dataset includes 51 structural connection matrices for the autistic people and 43 structural connection matrices for healthy people. The functional and structural networks of this data set can be downloaded from the website introduced in [36].

- **ABIDE Dataset**

In the Autism Brain Imaging Data Exchange (ABIDE) project, the functional and structural brain networks of autistic and healthy people have been collected from various laboratories. This dataset includes 539 preprocessed imaging data for affected people and 573 preprocessed imaging data for healthy people and five teams with their own tools [37] have performed the functional preprocessing. The samples related information and imaging parameters for each site in ABIDE are listed on the project's website[1]. To receive the brain images and time series from each region of the brain, a simple python script has been implemented utilizing nilearn library. The Harvard-Oxford atlas and the cpac preprocessing pipeline have been selected to fetch time series and a band-pass filter is applied to the resulting signals. Finally, the functional networks are obtained by the correlation method. The Harvard-Oxford atlas consists of 111 cortical and subcortical regions [38].

- **Tools**

In order to implement and simulate the proposed method, the python programming language and pycharm environment have been used. To implement the genetic evolutionary algorithm, we have utilized the DEAP (v1.3) evolutionary computational framework. To train and evaluate various machine-learning models we have used the Sklearn (v0.21), and finally to model the brain networks, the networkX library (v1.2) has been employed.

- **Brain Network Threshold**

Since there is a correlation (either negatively or positively) between each pair of regions in functional networks, the resulting graph is almost complete; in structural networks, due to the existence of a lot of physical connections (either strong or weak) between regions, the obtained raw network is almost complete too. Both structural and functional networks are weighted graphs. To appear the topological structure in the two group networks, it seems necessary to remove some unimportant edges with an appropriate threshold. In the present study, the value of density s has been determined as a constant value, and for each brain network, the threshold t is decided in such a way that the density of the network is equal to s. In the network, edges with lower weight than the threshold t are eliminated. The graph density $S(G)$ is obtained using Equation (1).

$$S(G) = \frac{2 * |E|}{n * (n-1)} \tag{1}$$

- **Graph Entropies**

Graph entropy is a criterion of the information theory to measure the complexity of a graph and describe the topological features of the graph. Following the introduction of entropy in the information and communication by Shannon, some generalizations of entropy calculations such as Renyi and Daroczy entropies were presented. Probability distribution entropy, in addition to being a criterion of uncertainty, is also a criterion of information. The amount of information we receive when we see the result of a test is numerically equal to the uncertainty of the test result before it is performed.

Studies of graph information content began in the late 1950s [39, 40]. After Shannon's famous article on entropy, a variety of entropies were defined for graphs, and these criteria well illustrated the graph properties. Dehmer and Mowshowitz applied the generalized entropy criteria to the graph in a new way [41]. Lee and et al. [42] introduced various entropies for each of the graph-related matrices, calling it a tool for describing the topological features of the graph. In this paper, for the first time, these criteria are utilized to describe the functional and structural networks of the brain. More details on these entropies can be seen in table 1 [42].

---

[1] http://fcon_1000.projects.nitrc.org/indi/abide/abide_I.html

Table 1: The entropies used in the current study

| Matrix | Entropies | Desc. |
|---|---|---|
| Adjacency Matrix | $I^1(G) = 1 - E(G)^{-2} \sum_{i=1}^{n} |\mu_i|^2$ <br> $I_\alpha^2(G) = \frac{1}{1-\alpha} \log \sum_{i=1}^{n} (\frac{|\mu_i|}{E(G)})^\alpha, \alpha \neq 1$ <br> $I_\alpha^3(G) = \frac{1}{2^{1-\alpha}-1} \sum_{i=1}^{n} (\frac{|\mu_i|}{E(G)})^\alpha - 1, \alpha \neq 1$ | $\mu_1, \mu_2, \ldots, \mu_n$ are eigenvalues of the adjacency matrix <br> $E(G) = \sum_{j=1}^{n} |\mu_j|$ |
| Unsigned Laplacian Matrix | $I_Q^1(G) = 1 - \frac{1}{4m^2}(M_1 + 2m)$ <br> $I_\alpha^2(G) = \frac{1}{1-\alpha} \log \frac{M_\alpha^*}{(2m)^\alpha}, \alpha \neq 1$ <br> $I_\alpha^3(G) = \frac{1}{2^{1-\alpha}-1} (\frac{M_\alpha^*}{(2m)^\alpha} - 1), \alpha \neq 1$ <br> $I_I^1(G) = 1 - \frac{2m}{IE^2(G)}$ <br> $I_\alpha^2(G) = \frac{1}{1-\alpha} \log \frac{M_\alpha^*}{IE^\alpha(G)}, \alpha \neq 1$ <br> $I_\alpha^3(G) = \frac{1}{2^{1-\alpha}-1} (\frac{M_\alpha^*}{IE^\alpha(G)} - 1), \alpha \neq 1$ | $q_1, q_2, \ldots, q_n$ are eigenvalues of the unsigned laplacian matrix <br> $M_1$ is the first Zagreb index <br> $\sum_{i=1}^{n} q_i = 2m$ <br> $M_\alpha^* = \sum_{i=1}^{n} |q_i|^\alpha$ <br><br> $q_1, q_2, \ldots, q_n$ are eigenvalues of the unsigned laplacian matrix <br> $\sum_{i=1}^{n} q_i = 2m$ <br> $M_\alpha^* = \sum_{i=1}^{n} \sqrt{q_i}^\alpha$ <br> $IE(G) = \sum_{i=1}^{n} \sqrt{q_i}$ |
| Unsigned Normalized Laplacian Matrix | $I_{\zeta(\varrho)}^1(G) = 1 - \frac{1}{n^2}(n + 2R_{-1}(G))$ <br> $I_\alpha^2(G) = \frac{1}{1-\alpha} \log \frac{M_\alpha^*}{n^\alpha}, \alpha \neq 1$ <br> $I_\alpha^3(G) = \frac{1}{2^{1-\alpha}-1} (\frac{M_\alpha^*}{n^\alpha} - 1), \alpha \neq 1$ | $\mu_1, \mu_2, \ldots, \mu_n$ are eigenvalues of the normalized Laplacian matrix <br> $q_1, q_2, \ldots, q_n$ are eigenvalues of the unsigned normalized Laplacian matrix <br> $\sum_{i=1}^{n} \mu_i = n$ <br> $R_\beta(G) = \sum_{uv \in E} (d_u d_v)^\beta$ <br> $M_\alpha^* = \sum_{i=1}^{n} |q_i|^\alpha$ |
| Distance Matrix | $I_D^1(G) = 1 - \frac{4}{DE^2(G)}(2WW(G) - W(G))$ <br> $I_\alpha^2(G) = \frac{1}{1-\alpha} \log \frac{M_\alpha^*}{DE^\alpha(G)}, \alpha \neq 1$ <br> $I_\alpha^3(G) = \frac{1}{2^{1-\alpha}-1} (\frac{M_\alpha^*}{DE^\alpha(G)} - 1), \alpha \neq 1$ | $\mu_1, \mu_2, \ldots, \mu_n$ are eigenvalues of the distance matrix <br> $DE(G) = \sum_{i=1}^{n} |\mu_i|$ <br> $W_k(G) = \sum_{1 \leq i < j \leq n} (d_{ij})^k$ $\quad WW(G) = \frac{1}{2}(W_1(G) + W_2(G))$ <br> $M_\alpha^* = \sum_{i=1}^{n} |\mu_i|^\alpha$ |
| Randic Adjacency Matrix | $I_R^1(G) = 1 - \frac{2}{RE^2(G)} R_{-1}(G)$ <br> $I_\alpha^2(G) = \frac{1}{1-\alpha} \log \frac{M_\alpha^*}{RE^\alpha(G)}, \alpha \neq 1$ <br> $I_\alpha^3(G) = \frac{1}{2^{1-\alpha}-1} (\frac{M_\alpha^*}{RE^\alpha(G)} - 1), \alpha \neq 1$ | $\rho_1, \rho_2, \ldots, \rho_n$ are eigenvalues of the randic adjacency matrix <br> $RE(G) = \sum_{i=1}^{n} |\rho_i|$ <br> $R_\beta(G) = \sum_{uv \in E} (d_u d_v)^\beta$ <br> $M_\alpha^* = \sum_{i=1}^{n} |\rho_i|^\alpha$ |
| Randic Incidence Matrix | $I_{I_R}^1(G) = 1 - \frac{r}{I_R E^2(G)}$ <br> $I_\alpha^2(G) = \frac{1}{1-\alpha} \log \frac{M_\alpha^*}{I_R E^\alpha(G)}, \alpha \neq 1$ <br> $I_\alpha^3(G) = \frac{1}{2^{1-\alpha}-1} (\frac{M_\alpha^*}{I_R E^\alpha(G)} - 1), \alpha \neq 1$ | $\sigma_1, \sigma_2, \ldots, \sigma_n$ are eigenvalues of the randic incidence matrix <br> $r$ is number of no isolated vertices <br> $I_R E(G) = \sum_{i=1}^{n} \sigma_i$ <br> $M_\alpha^* = \sum_{i=1}^{n} |\rho_i|^\alpha$ |
| General Randic Matrix | $I_{R_\beta}^1(G) = 1 - \frac{2}{RE_\beta^2(G)} R_{2\beta}(G)$ <br> $I_\alpha^2(G) = \frac{1}{1-\alpha} \log \frac{M_\alpha^*}{RE_\beta^\alpha(G)}, \alpha \neq 1$ <br> $I_\alpha^3(G) = \frac{1}{2^{1-\alpha}-1} (\frac{M_\alpha^*}{RE_\beta^\alpha(G)} - 1), \alpha \neq 1$ | $\gamma_1, \gamma_2, \ldots, \gamma_n$ are eigenvalues of the general randic matrix <br> $RE_\beta(G) = \sum_{i=1}^{n} |\gamma_i|$ <br> $R_\beta(G) = \sum_{uv \in E} (d_u d_v)^\beta$ <br> $M_\alpha^* = \sum_{i=1}^{n} |\gamma_i|^\alpha$ |

- **Finding sub-network with Genetic Algorithm**

In the current study, for each functional or structural dataset, we are looking for a discriminative sub-network which indicates differences between healthy and autism groups better. Vividly, the extracted sub-network has been used to performing the classification task instead of the whole brain network, with the aim of autism screening. Sub-network classification leads to improvement of the interpretability of the test results in addition to increasing screening accuracy.

In the current study, for the first time, we adopted a genetic algorithm to extract a discriminative sub-network. The first step in implementing a genetic algorithm is to choose how genotypes (or problem solutions) are represented. Due to the nature of the problem, the binary representation method has been used.

The length of genotypes varies due to the difference in the number of nodes in the brain networks. In the UCLA dataset, each network has 256 nodes since the power brain atlas contains 256 regions, so the length of the genotype is 256. Also in the ABIDE dataset, because of adopting Harvard-Oxford Atlas, the length of the genotypes is 111. Each element of the solution (genotype) string indicates the presence or absence of a node in the brain network. Being one or zero an element in genotype means that a node exists or not in the brain network. When a node does not exist in the network, all its adjacent edges eliminated too. One of the most important steps in the genetic algorithm is adopting an appropriate fitness function. In the present study, since each genotype in the real space is a brain sub-network, the fitness value is the accuracy of the classification of brain networks into two groups using a genotype. For fitness evaluation, a machine-learning model is trained using the entropy features and then evaluated using the cross-validation method. The average classification accuracy determines the fitness of the genotype.

The initial population consists of 100 randomly generated genotypes. In the process of the genetic algorithm, the mutation and single-point and two-point cross-over genetic operators have been used. In the proposed method, the stop condition of the genetic algorithm depends on the number of generations, the maximum and average of population fitness, and finally on the population diversity (sum of Hamming distances between each pair of genotypes). In the selection phase, the truncated rank based selection method is used. The values of the genetic algorithm parameters are listed in Table 2.

**Table 2: The Values of the Genetic Algorithm Parameters**

| Parameter | value |
| --- | --- |
| Number of primary population members | 100 |
| Probability of combining two solutions | 0.30 |
| Probability of mutation of a solution | 0.30 |
| Probability of mutation of each element of the solution | 0.05 |

For each dataset (UCLA functional, UCLA structural and ABIDE functional) and machine-learning models (k nearest neighbor, decision tree, support vector machine, and logistic regression as well as ensemble methods such as AdaBoost, and bagging decision tree) the genetic algorithm has been executed 20 times with different initial populations. Consequently, the result (the most appropriate sub-network) has been reported among all the executions for each dataset. In Figure 2 the extracted discriminative nodes are shown for UCLA functional / structural datasets.

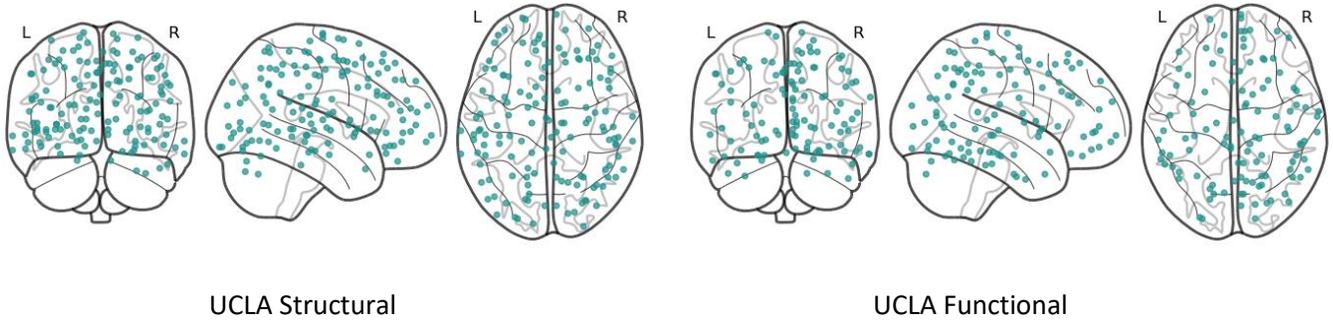

UCLA Structural                                UCLA Functional

Figure 2: The extracted discriminative nodes for UCLA functional / structural dataset

## 4- Simulation Results

The following are the results of simulating the classification of the functional and structural brain networks into two categories: healthy controls and autism disorders. In the structural networks of the UCLA dataset, the most discriminative sub-network has been obtained using the genetic algorithm with the fitness function of the SVM classification accuracy. The generation graph related to the best evolution among evolutions with different initial populations can be seen in Figure 3. The maximum global fitness in each generation is the best classification accuracy obtained by the genetic algorithm up to that generation, and the average and minimum local fitness has been obtained between population members in each generation.

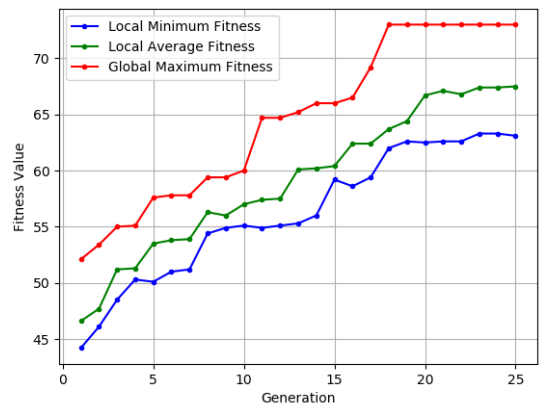

**Figure 3: Global maximum fitness and minimum/average local fitness for each generation in the best execution, for structural networks of the UCLA dataset**

For functional networks of the UCLA dataset, the most discriminative sub-network is obtained using the genetic algorithm with the Adaboost classification accuracy fitness function, and the best related generation graph can be seen in Figure 4 among different initial populations.

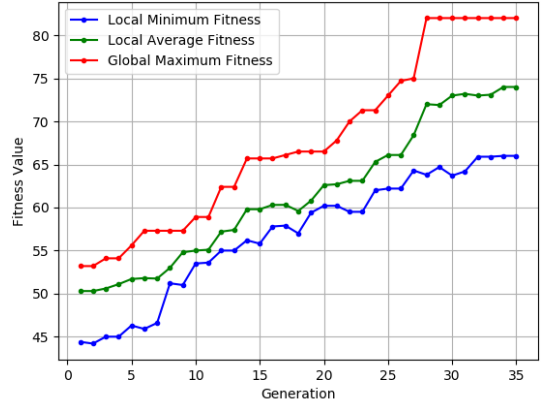

**Figure 4: Global maximum fitness and minimum/average local fitness for each generation in the best execution, for functional networks of the UCLA dataset**

For the functional networks of the ABIDE dataset, the most discriminative sub-network is obtained using the genetic algorithm with the fitness function of the KNN model classification accuracy. The generation graph related to the best performance among the different initial populations can be seen in Figure 5.

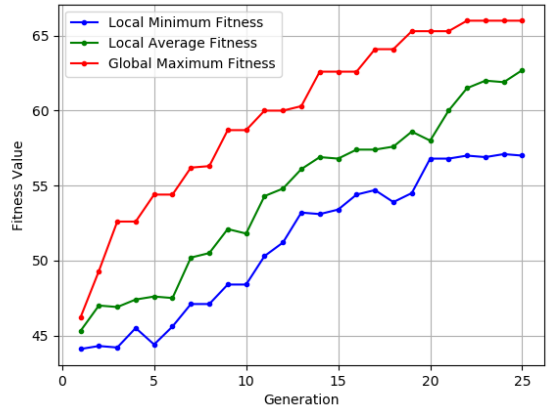

**Figure 5: Global maximum fitness and minimum/average local fitness for each generation in the best execution, for functional networks of the ABIDE dataset**

In Table 3, the best classification results are reported using the discriminated sub-networks derived from the genetic algorithm for each dataset. The detailed reports on the best classification can be seen in Table 4.

**Table 3: Best performances of the genetic algorithm using various machine-learning models In fitness calculation**

| Classification | Functional ABIDE | Functional UCLA | Structural UCLA |
|---|---|---|---|
| | Accuracy | Accuracy | Accuracy |
| KNN | **66.1** | 55.0 | 61.0 |
| SVM | 64.2 | 61.4 | **73.1** |
| DT | 62.6 | 64.0 | 62.5 |
| LR | 65.0 | 63.5 | 66.3 |
| Adaboost | 64.7 | **82.2** | 66.1 |
| Bagging | 61.0 | 0.62 | 60.0 |
| **Max** | **66.1** | **82.2** | **73.1** |

**Table 4: Detailed evaluation results of the best genetic algorithm execution**

| | Accuracy | Precision | Recall |
|---|---|---|---|
| UCLA Structural | 73.1 | 71.1 | 89.0 |
| UCLA Functional | 82.2 | 81.0 | 85.0 |
| ABIDE Functional | 66.1 | 67.5 | 61.2 |

Valuable efforts have been made to screen autism disorder using brain structural network. In the current study, we have compared our method performance with the best researches in recent years. The evaluation results indicate that the proposed method in terms of classification accuracy, is 9% more accurate than Petrov [26] and Dodonova [30] and 5% than Tolan [24] and Dodero [25].

In the field of autism screening using brain networks, more efforts have been made in functional networks than in the structural networks due to the existence of more datasets for functional networks. In UCLA functional dataset the results indicate the improved performance of our proposed method in the classification than other efforts. The results of the classification show that the classification accuracy of the extracted sub-network compared to Eslami research [31], 9%, with Atai research [29], 13%, with Tolan research [24], 14% and with Dodero research [25], 21% has been improved.

The comparison between the proposed method and the method [28] on the ABIDE functional dataset with common brain atlas demonstrates a 4% improvement in the accuracy of the classification for autism screening. The mentioned results can be seen in Table 5.

**Table 5: Comparison of the classification accuracy between the proposed method and previous researches**

| Dataset | The Proposed Method Accuracy (%) | Method | Accuracy (%) |
|---|---|---|---|
| **UCLA Structural** | **73** | E. Tolan et al.[24] | 68 |
| | | L. Dodero et al.[25] | 68 |
| | | D. Petrov et al.[26] | 64 |
| | | Y. Dodonova et al.[30] | 64 |
| **UCLA Functional** | **82** | T. Eslami et al.[31] | 73 |
| | | E. Tolan et al.[24] | 68 |
| | | L. Dodero et al. [25] | 61 |
| | | S. Ataei et al. [29] | 69 |
| **ABIDE Functional** | **66** | S. I. Ktena et al.[28] | 62 |

## 5- Conclusion and Future Works

In this paper, for the first time, the evolutionary algorithms have been utilized to extract differentiating sub-networks in the brain networks. Extracting these sub-networks, in addition to improving the accuracy of classification, increases the interpretability of the study results. The proposed method presented has been evaluated by the valid datasets and the results demonstrate that this method has better performance to screen the autism disorder than other methods.

For UCLA structural dataset, the genetic algorithm with an SVM accuracy fitness evaluation has been found a sub-network that detects autistic people with 73% accuracy and performs 5% better than the best research on this dataset. For UCLA functional dataset, the genetic algorithm with an Adaboost accuracy fitness evaluation has reached a sub-network that has the ability to detect autistic people with an average accuracy of 82% and is 13% more accurate than the best research on this dataset. For ABIDE functional dataset, the genetic algorithm with a KNN accuracy fitness evaluation has extracted a sub-network that accurately detects 66% of autistic and healthy people and has been improved by 4% compared to the most significant previous research for this dataset with common brain atlas.

In addition to the general entropies mentioned in this paper, there are other types of entropies that have not been studied in the brain networks, and research on these entropies can be considered as future work. Moreover, studying the different types of network energy and their relationship to the network entropies can help machine learning models in classification and improve screening accuracy.